\documentclass[10pt,twocolumn,letterpaper]{article}

\usepackage{cvpr2016pkg/cvpr}

\usepackage{times}
\usepackage{epsfig}
\usepackage{graphicx}
\usepackage{amsmath}
\usepackage{amssymb}
\usepackage{booktabs}
\usepackage{latex_pkgs/msmaths}
\usepackage{latex_pkgs/msformatting}
\usepackage[conf]{optional}

\usepackage{caption}
\usepackage{cuted}
\usepackage{fancyhdr,floatpag}

\usepackage[pagebackref=true,breaklinks=true,letterpaper=true,colorlinks,bookmarks=false]{hyperref}

\cvprfinalcopy % *** Uncomment this line for the final submission

\def\cvprPaperID{2441} % *** Enter the CVPR Paper ID here

\ifcvprfinal\fi
\begin{document}

%%%%%%%%% TITLE
\title{
  Untrimmed Video Classification for Activity Detection: \\
  submission to ActivityNet Challenge
}

\author{Gurkirt Singh \qquad Fabio Cuzzolin \\ Artificial Intelligence and Vision research group \\ Oxford Brookes University\\ {\tt\small \{15056568,fabio.cuzzolin\}@brookes.ac.uk}}

% \author{Gurkirt Singh 
% 
% % For a paper whose authors are all at the same institution,
% % omit the following lines up until the closing ``}''.
% % Additional authors and addresses can be added with ``\and'',
% % just like the second author.
% % To save space, use either the email address or home page, not both
% \and
% Fabio Cuzzolin
% 
% Institution2\\
% First line of institution2 address\\
% {\tt\small secondauthor@i2.org}
% }

\maketitle
%\thispagestyle{empty}
% \begin{strip}
%   \centering
%   \includegraphics[width=0.95\textwidth]{figures/introductionTeaser.pdf}
%   \vskip 0.1cm
% 
%   \begin{minipage}[adjusting]{0.95\textwidth}
%     Figure 1: 
%     A video sequence taken from the LIRIS-HARL dataset plotted in space-and time.
%     \textbf{(a)} A top down view of the video plotted with the detected action tubes of class `handshaking' in green, and `person leaves baggage unattended' in red. Each action is located to be within a space-time tube.
%     \textbf{(b)} A side view of the same space-time detections. Note that no action is detected at the beginning of the video when there is human motion present in the video.
%     \textbf{(c)} The detection and instance segmentation result of two actions occurring simultaneously in a single frame.
%   \end{minipage}
%   \label{fig:introductionTeaser}
% \end{strip}
% \setcounter{figure}{1}

%%%%%%%%% ABSTRACT
\begin{abstract}  
  Current state-of-the-art human activity recognition is focused on the classification of temporally trimmed videos in which only one action occurs per frame. 
We propose a simple, yet effective, method for the temporal detection of activities in temporally untrimmed videos
with the help of untrimmed classification.
Firstly, our model predicts the top $k$ labels for each untrimmed video by analysing global video-level features. 
Secondly, frame-level binary classification is combined with dynamic programming to generate the temporally trimmed \emph{activity proposals}.
Finally, each proposal is assigned a label based on the global label, and scored with the score of the temporal activity proposal and the global score. % ?? unclear
Ultimately, we show that untrimmed video classification models can be used as stepping stone for temporal detection. 
Our method wins runner-up prize in ActivtiyNet Detection challenge 2016. 

\end{abstract}
%%%%%%%%% BODY TEXT
\vspace{-8mm}
\section{Introduction}
% motivation: real-world scenarios
Emerging real-world applications require an all-round approach to the machine understanding of human behaviour, which goes beyond the recognition of simple, isolated activities from video.

%Existing works on action recognition have achieved impressive recognition rates, however they are mostly focused on 
%action classification~\cite{laptev-2008,wang-2011,wang-2013, Shuiwang-2013,Karpathy-2014,Simonyan-2014} and localisation~\cite{Georgia-2015a, Weinzaepfel-2015}.
As a step towards this ambitious goal, in this work we address the problem of detecting the temporal bounds of activities in temporally untrimmed videos.

\section{Methodology} \label{sec:methodology}
%% Introduction
Whereas (i) video-level features are used for untrimmed video classification task, (ii) frame-level features are used for activity proposal generation and scoring. 
Finally, (iii) a video's classification score is augmented with the scores of the activity proposals for proposal classification.

\subsection{Features}
We make use of the features provided on ActivityNet's~\cite{caba2015activitynet} web page\footnote{http://activity-net.org/challenges/2016/download.html}. 

\subsubsection{Video-level features}

\textbf{ImageNetShuffle} features are video-level features generated by~\cite{MettesICMR16} using a Google inception net (GoogLeNet~\cite{szegedy2015going}). 
CNN features are extracted from the pool5 layer of GoogLeNet~\cite{szegedy2015going} at a two frames per second rate.
Frame-level CNN features are mean pooled to construct a representation for the whole video. Mean pooling is followed by L1-normalisation.

We train a one-versus-rest linear SVM for each class,
and use the resulting SVM scores $S^i = \{s_{1}^{i},...,s_{c}^{i},...s_{C}^{i}\}$, where $C$ is number of classes, as INS features. % are these scores *the* INS features, are just part of them?

\textbf{Motion Boundary Histogram (MBH) features} are generated with the aid of the 
improved trajectories~\cite{wang-2013} executable\footnote{http://lear.inrialpes.fr/people/wang/improved\_trajectories}.
We train another battery of one-versus-rest SVMs using a linear kernel on the MBH features,
and use the resulting SVM scores $S^m = \{{s_{1}^{m},...,s_{c}^{m},...s_{C}^{m}}\}$ as global video features.

\subsubsection{Frame level features}

\textbf{C3D Features features} are generated at 2 frames per second using a C3D network~\cite{tran2014learningC3D} with temporal resolution of 16 frames.
Once again we train a frame level one-versus-rest SVM classifier for each activity class using a linear kernel. 
The scoring of frame $t$ is defined by the resulting SVM scores: $S^3_t = \{{s_{1}^{3},...,s_{c}^{3},...s_{C}^{3}}\}$.
Finally, we perform mean pooling along the frames for each class to get another score vector $S^3$, which is used for video classification.

\subsection{Untrimmed video classification} \label{sec:untrimmed-classification}

Untrimmed video classification is achieved by fusing all video level scores using a linear SVM as a meta classifier. 
Video level scores ($S^i$, $S^m$ and $S^3$) are stacked up to make a single score vector.
A linear SVM is trained on the training set of stacked scores, and evaluated on the validation and testing sets.
The output scores $S^s$ outputted by the meta SVM are normalised by dividing them by the sum of the top $k$ scores. 
The parameter $k$ was cross-validated on the validation set and set to 3 -- it contributes to improve the mean average precision metric. 

We believe that, since SVM scores are not probabilities, normalisation by top $k$ scores is required to be able to compare them across all videos.

\subsection{Activity detection in untrimmed videos}

Activity proposals are detected by (i) training a binary random forest (RF) classifier~\cite{breiman2001random} for each class on the frame-level C3D features, and (ii) casting activity proposal generation as an optimisation problem~\cite{Evangel-2014}, which makes use of these binary decisions.

\subsubsection{Binary random forest classification}

The binary RF classifies each frame into a negative (i.e. no activity taking place) or a positive bin (i.e. something is happening).
%We chose random forest because it's probabilities nature. 
The positive score of a frame $t$  is denoted by $s^r_t$.
Temporal trimming is then achieved by dynamic programming as follows.

\subsubsection{Activity proposal generation} \label{sec:proposal-generation}

Given the frame-level scores $\{s^r_t, t=1,...,T\}$ for a video of length $T$, we want to assign to each frame a binary label \mbox{${l}_{t}$ $\in$ $\{1,0\}$}
(where zero represents the `background' or `no-activity' class),
which maximises:
\begin{equation}
\label{eqn:secondpassenergy}
E(L) =  \sum_{t=1}^T s^r_t - \lambda \sum_{t=2}^T \psi_l \left( \Scalar{l}_{t}, 
\Scalar{l}_{t-1} \right) ,
\end{equation}
%where $\Vector{L}_p~=~(l_1,l_2,l_3, \dots l_T)$, is a sequence of $0$ and $1$ labels for a path $p$, and 
where 
$\lambda$ is a scalar parameter, 
%$s_{l_t=1}(\Vector{b}_t) = s^*_c(\Vector{b}_t)$, $s_{l_t=0}(\Vector{b}_t) = 1 - s^*_c(\Vector{b}_t)$, and $s^*_c(\Vector{b}_t)$ are the augmented scores (\ref{eqn:appflow_fusion}).\\
and the pairwise potential $\psi_l$ is defined as: 

$\psi_l(l_t,l_{t-1}) = 0$ if $l_t=l_{t-1}$, $\psi_l(l_t,l_{t-1}) =\alpha$ otherwise,

(where $\alpha$ is a parameter which we set by cross validation).
This penalises labellings $L = \{l_1,...,l_T\}$ which are not smooth, thus enforcing a piecewise constant solution.
All contiguous sub-sequences  form the desired activity proposal (which can be as many as there are instances of activities). 
Each activity proposal is assigned a global score $S_a$ equal to the mean of the scores of its constituting frames. 
This optimisation problem can be efficiently solved by dynamic programming~\cite{Evangel-2014}. 
It can easily be extended for simultaneous detection and classification~\cite{Evangel-2014}.

\subsubsection{Activity detection}

%We take top activity classification labels and scores to label activity proposals. 
The top (in this case 2) activity proposals in each video are assigned the label of top untrimmed classification class (\S \ref{sec:untrimmed-classification}).
For example, if $c={10}$ is the top class for the video with score $S_{10}^s$, and $a$ is the top activity proposal with score $S_a$ (\S \ref{sec:proposal-generation}),
then a detection of class $10$ is flagged with the temporal bounds determined by activity proposal $a$ and score $S_{10}^a = S_{10}^s * S_a$.
Similarly, we can generate more detections for each of the top classes by using top activity proposals. % ?? unclear

\floatpagestyle{fancy}
\section{Implementation}

We used the precomputed features provided by the competition organisers. We used SciKit-learn %http://scikit-learn.org/stable/}
for linear SVM and random forest Implementation.  
Our code available at \url{https://github.com/gurkirt/actNet-inAct}.

\section{Results}

We report results for untrimmed classification and activity detection on ActivityNet~\cite{caba2015activitynet}. 
We use the same evaluation setting as described in challenge~\cite{caba2015activitynet}.

\subsection{Untrimmed classification}

\begin{table}[ht!]
\begin{center}
\resizebox{0.45\textwidth}{!}{ 
\begin{tabular}{l|ccc|ccc}
\hline % check my corrections here are correct
& \multicolumn{3}{c}{Validation Set} & \multicolumn{3}{c}{Testing Set} \\
\hline
Method &TOP-1&TOP-3&mAP&TOP-1&TOP-3&mAP\\
\hline
Caba \etal~\cite{caba2015activitynet} & - & - & 42.50\% & - & - & 42.20\% \\
proposed & 76.89\% & 89.25\% & 81.99\% & 77.08\% & 89.38\% & 82.49\% \\
\hline
\end{tabular}
}
\end{center}
\vspace{-4mm}
\caption{Untrimmed classification performance on validation and testing set in percentage.}
\label{table:untrimmed}
\end{table}

\subsection{Activity detection}

\begin{table}[ht!]
\begin{center}
\resizebox{0.45\textwidth}{!}{ 
\begin{tabular}{l|ccc|ccc}
\hline
TIoU threshold $\delta$ = & 0.1 & 0.2 & 0.3 & 0.4 & 0.5 \\
\hline
Validation-Set Caba \etal~\cite{caba2015activitynet}& 12.50\% & 11.90\% & 11.1\% & 10.40\% & 09.70\% \\
Validation-Set proposed  & 52.12\% & 47.94\% & 43.50\% & 39.22\% & 34.47\% \\
\hline
Testing-Set proposed  & - & - & - & - & 36.40\% \\ % why are results not reported for other threshold values?
\hline
\end{tabular}
}
\end{center}
\vspace{-4mm}
\caption{Activity detection  performance on validation and testing set. Quantity $\delta$ is the Temporal Intersection over Union  (TIoU) threshold.}
\label{table:detection}
\end{table}

\begin{strip}
%\twocolumn[{
\begin{center}
    \centering
    \includegraphics[width=1.0\textwidth]{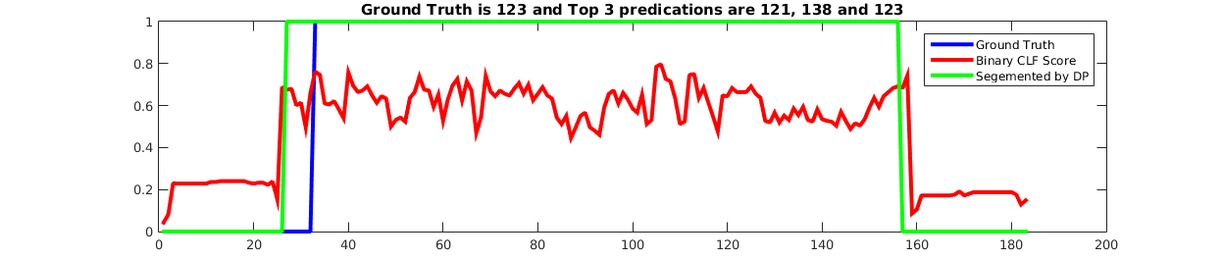}
    \vspace{-7mm}
    \captionof{figure}{Plot shows ground truth in blue, binary classifier score in red and piece-wise constant proposal produce by DP optimisation. 
    Binary classifier scores are high where activity is happening, which produces well aligned segment proposal with ground truth.}
    \vspace{-16mm}
\end{center}%
%}]
\end{strip}
\setcounter{figure}{1}

\begin{strip}
%\twocolumn[{
\begin{center}
    \centering
    \includegraphics[width=1.0\textwidth]{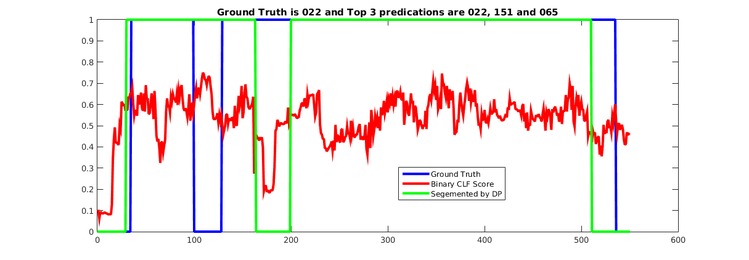}
    \vspace{-8mm}
    \captionof{figure}{Plot shows ground truth in blue, binary classifier score in red and piece-wise constant proposal produce by DP optimisation. 
    We can see two proposal are mostly aligned with ground truth. }
    \vspace{-16mm}
\end{center}
%}]
\end{strip}

\begin{strip}
%\twocolumn[{
\begin{center}
    \centering
    \includegraphics[width=1.0\textwidth]{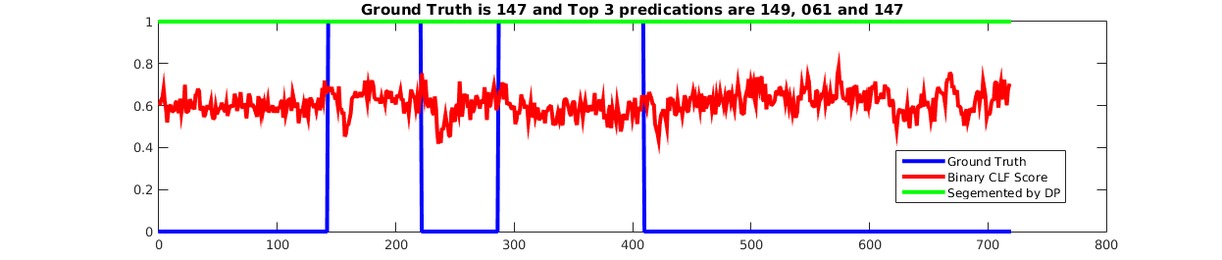}
    \vspace{-6mm}
    \captionof{figure}{Plot shows ground truth in blue, binary classifier score in red and piece-wise constant proposal produce by DP optimisation. 
    Our method completely fails to trim two instance but only produce one segment for whole video as binary classifier score are high throughout the video duration.}
    \vspace{-4mm}
\end{center}%
%}]
\end{strip}

\section{Conclusion and Future Work}

We show that activity detection can be achieved via untrimmed video classification. 
Our dynamic programming-based approach is efficient, and has shown a clear potential for generating good quality activity proposal. 

The approach can be easily extended for simultaneous detection and classification without requiring classification scores at video level, 
which open ups the opportunity for online activity classification, detection and prediction. 

{\small
\bibliographystyle{cvpr2016pkg/ieee}
\bibliography{refcvpr2016}

\begin{thebibliography}{1}\itemsep=-1pt

\bibitem{breiman2001random}
L.~Breiman.
\newblock Random forests.
\newblock {\em Machine learning}, 45(1):5--32, 2001.

\bibitem{caba2015activitynet}
F.~Caba~Heilbron, V.~Escorcia, B.~Ghanem, and J.~Carlos~Niebles.
\newblock Activitynet: A large-scale video benchmark for human activity
  understanding.
\newblock In {\em Proceedings of the IEEE Conference on Computer Vision and
  Pattern Recognition}, pages 961--970, 2015.

\bibitem{Evangel-2014}
G.~Evangelidis, G.~Singh, and R.~Horaud.
\newblock Continuous gesture recognition from articulated poses.
\newblock In {\em ECCV Workshops}, 2014.

\bibitem{MettesICMR16}
P.~Mettes, D.~Koelma, and C.~G.~M. Snoek.
\newblock The imagenet shuffle: Reorganized pre-training for video event
  detection.
\newblock In {\em Proceedings of the {ACM} International Conference on
  Multimedia Retrieval}, New York, USA, 2016.

\bibitem{szegedy2015going}
C.~Szegedy, W.~Liu, Y.~Jia, P.~Sermanet, S.~Reed, D.~Anguelov, D.~Erhan,
  V.~Vanhoucke, and A.~Rabinovich.
\newblock Going deeper with convolutions.
\newblock In {\em Proceedings of the IEEE Conference on Computer Vision and
  Pattern Recognition}, pages 1--9, 2015.

\bibitem{tran2014learningC3D}
D.~Tran, L.~Bourdev, R.~Fergus, L.~Torresani, and M.~Paluri.
\newblock Learning spatiotemporal features with 3d convolutional networks.
\newblock {\em arXiv preprint arXiv:1412.0767}, 2014.

\bibitem{wang-2013}
H.~Wang and C.~Schmid.
\newblock {Action Recognition with Improved Trajectories}.
\newblock In {\em Proc. Int. Conf. Computer Vision}, pages 3551--3558, 2013.

\end{thebibliography}
}

\end{document}